\documentclass[11pt]{article}

\usepackage[final]{acl}

\usepackage{times}
\usepackage{latexsym}

\usepackage[T1]{fontenc}
\usepackage[utf8]{inputenc}

\usepackage{microtype}

\usepackage{float}
\usepackage{url}
\usepackage{amsmath,amssymb,amsfonts}
\usepackage{graphicx}
\usepackage{textcomp}
\usepackage{xcolor}
\usepackage{subcaption}
\usepackage{booktabs}
\usepackage{siunitx}
\title{Qomhrá: A Bilingual Irish and English Large Language Model}

\author{
  \begin{tabular}{ccc}
    \textbf{Joseph McInerney}$^{1,3}$ & \textbf{Khanh-Tung Tran}$^{2}$ & \textbf{Liam Lonergan}$^{1}$ \\
    \small\texttt{mcinerjo@tcd.ie} & \small\texttt{123128577@umail.ucc.ie} & \small\texttt{llonerga@tcd.ie} \\[3ex]  
    \textbf{Ailbhe Ní Chasaide}$^{1}$ & \textbf{Neasa Ní Chiaráin}$^{1}$ & \textbf{Barry Devereux}$^{3}$ \\
    \small\texttt{anichsid@tcd.ie} & \small\texttt{nichiarn@tcd.ie} & \small\texttt{b.devereux@qub.ac.uk}
  \end{tabular}
  \vspace{1em} \\ 
  \small
  $^1$Trinity College Dublin, $^2$University College Cork, $^3$Queen's University Belfast
}
\begin{document}
\maketitle

\begin{abstract}
Large language model (LLM) research and development has overwhelmingly focused on the world's major languages, leading to under-representation of low-resource languages such as Irish. This paper introduces \textbf{Qomhrá}, a bilingual Irish and English LLM, developed under extremely low-resource constraints. A complete pipeline is outlined spanning bilingual continued pre-training, instruction tuning, and the synthesis of human preference data for future alignment training. We focus on the lack of scalable methods to create human preference data by proposing a novel method to synthesise such data by prompting an LLM to generate ``accepted'' and ``rejected'' responses, which we validate as aligning with L1 Irish speakers.
To select an LLM for synthesis, we evaluate the top closed-weight LLMs for Irish language generation performance. Gemini-2.5-Pro is ranked highest by L1 and L2 Irish-speakers, diverging from LLM-as-a-judge ratings, indicating a misalignment between current LLMs and the Irish-language community. Subsequently, we leverage Gemini-2.5-Pro to translate a large scale English-language instruction tuning dataset to Irish and to synthesise a first-of-its-kind Irish-language human preference dataset. We comprehensively evaluate Qomhrá across several benchmarks, testing translation, gender understanding, topic identification, and world knowledge; these evaluations show gains of up to 29\% in Irish and 44\% in English compared to the existing open-source Irish LLM baseline, UCCIX. The results of our framework provide insight and guidance to developing LLMs for both Irish and other low-resource languages.  
\end{abstract}

\section{Introduction}
Whilst progress has been made in speech synthesis and recognition for Irish \cite{lonergan-etal-2022-automatic}, Irish remains the least-supported official European language in terms of language technology \cite{ele2022irish}. To address this, we present \textbf{Qomhrá}, an open-weights bilingual Irish and English LLM, developed under extremely low-resource constraints. Qomhrá supports the subsidiary development of chatbots and other NLP tools, applicable across education, translation, and public services, enabling access to language technologies for the Irish-language community. Crucially, Qomhrá fosters technological sovereignty for the Irish language by reducing reliance on proprietary API-based models. This mitigates risks associated with cost volatility and data privacy, while ensuring the community retains control over the adaptation of models to specific use cases.

\begin{figure}[t!]
    \centering
    \includegraphics[width=\linewidth]{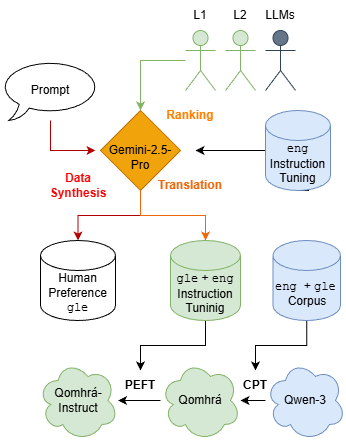}
    \caption{A High-Level Pipeline Overview of Data Synthesis, Model Ranking and LLM Training of the Qomhrá Irish-English language model.}
    \label{Pipeline Overview}
\end{figure}

Definitions of \textit{low-resource} vary from the socio-political to human expertise and data availability \cite{nigatu-etal-2024-zenos}. In the context of LLM development, we classify Irish as a low-resource language due to data scarcity, or, more precisely, the lack of permissively-licensed labelled data for instruction tuning and alignment. This data requires explicit human annotation, making it particularly valuable for training and evaluating LLMs but also it is typically costly to create.

To facilitate instruction tuning, we curate a labelled dataset by translating from English with a closed-weight model. To select this translator LLM, we evaluate budget and flagship models from top providers. We use three separate judges: an L1 speaker, an L2 speaker and the aggregate of three LLMs. 

An L2 speaker is considered in this context because they drastically outnumber L1 speakers of Irish. As of the 2022 census of Ireland \cite{cso2022-education-irish}, 40\% of people indicated that they speak Irish, of which, only 10\% indicated that they speak ``very well''. Therefore, if L2 speakers prove effective annotators, the cost of annotation can be reduced drastically. The same reasoning is applied for the inclusion of LLMs for annotation.

The development of Qomhrá further addresses labelled Irish data scarcity through the proposal of a novel synthesis framework that prompts the strongest available closed-weight model to translate an instruction-response pair from the high-resource language, English (eng), to the low-resource language, Irish (gle). The instruction remains constant but the LLM is prompted to generate a poor-quality and high-quality response mimicking human preference.  
This synthesis framework is applicable to other extreme low-resource language scenarios, where little is known about existing closed-weight LLM capabilities and annotated data is scarce. The framework is visualised in Fig \ref{Pipeline Overview} and investigates three key questions:
\begin{enumerate}
    \item To what degree does bilingual continued pre-training (CPT) improve Irish performance without diminishing existing English capabilities? 
    \item Accounting for API cost, which closed-weight LLM exhibits optimal Irish language chatbot performance as judged by an L1 Irish speaker?
    \item To what degree does synthetic preference data generated by SOTA LLMs align with human judgments? 
\end{enumerate}

The scope is bound by compute, data, and L1 speaker access. We train at the 8B scale without tokeniser retraining, and our evaluation relies on one L1 speaker and one L2 speaker. These restrictions are typical of low-resource research but still enable key methodological insights. \textbf{The contributions of our paper are five-fold:}
\begin{enumerate}
    \item The release of Qomhrá, an 8B parameter open-weight Irish and English LLM 
    \footnote{\url{https://huggingface.co/jmcinern/Qomhra}}, including a language-aware quantized version
    \footnote{\url{https://huggingface.co/jmcinern/Qomhra-AWQ}}.
    \item A first-of-its-kind, human evaluation and ranking of the top available closed-weight LLMs for Irish.
    \item A 30K sample parallel English-Irish instruction tuning dataset that significantly improves the LLM's ability to follow instructions
    \footnote{\url{https://huggingface.co/datasets/jmcinern/Dolly-V2-gle}}.
    \item A novel method to synthesise human preference data by prompting an LLM to generate ``accepted'' and ``rejected'' translations of an existing English prompt-response pair.
     \item We release a 1K sample Irish human preference dataset that is shown to align with an L1 Irish speaker
     \footnote{\url{https://huggingface.co/datasets/jmcinern/LIMA-gle-DPO}}.
\end{enumerate}

\section{Related Work}

\subsection{LLMs for Low-Resource Languages}
While major languages, particularly English, dominate interest in state-of-the-art LLM performance, there have also been efforts that have focused on adapting models to low-resource languages. Initial efforts for Irish focused on encoder-only models such as gaBERT and gaELECTRA \cite{barry-etal-2022-gaBERT}. More recently, the focus has shifted to generative decoder models, such as UCCIX for Irish \cite{tran2024uccixirishexcellencelargelanguage} and elsewhere, Latxa for Basque \cite{etxaniz-etal-2024-latxa}, adapting to their respective low-resource languages via continued pre-training \cite{gururangan-etal-2020-CPT}. We note that both efforts report cases of catastrophic forgetting \cite{cata_MCCLOSKEY1989109}, where English performance declines, and thus we include a significantly higher proportion of English training data for bilingual capabilities.

\subsection{Instruction Tuning}
For high-resource languages, instructions annotated by humans are abundant. In low-resource settings, they are often either low-quality or absent. A prevailing strategy is \textit{translate-train} \cite{conneau-etal-2018-xnli}, where data is translated from source to target language and then used to train the model. This is effective at augmenting low-resource instruction tuning data \cite{singh2024ayadatasetopenaccesscollection}.

We adopt this methodology by translating the Dolly V2 dataset \cite{DatabricksBlog2023DollyV2}, but we extend the pipeline by evaluating the translator model first, since in low-resource settings model providers often do not evaluate low-resource languages. This extension helps others looking to select models for Irish data synthesis and highlights a key step for others seeking to synthesise data in other low-resource languages. 
\subsection{Alignment}
After instruction following, the next step is to align models to human preferences, which has typically relied on Reinforcement Learning from Human Feedback (RLHF) \cite{Christiano_1017_RLHF}, a procedure which is computationally complex and costly. Direct Preference Optimization (DPO) \cite{DPO_Rafilov_2023} offers a more stable, data-efficient alternative but requires paired ``chosen'' and ``rejected'' response data, which does not exist for Irish. Synthetic data generation for DPO has been explored for other languages \cite{kazakh-pseudo-DPO}, often using a teacher model to score responses. We propose a simplified, novel approach that explicitly prompts a strong multilingual teacher (Gemini-2.5-Pro) to synthesise the preference signal itself by generating contrasting translation qualities, thereby bypassing the need for a fine-tuned teacher model.

\section{Model Training Framework}
\subsection{Continued Pre-training Data}
As with the GaBERT and UCCIX models, we use the available web-crawled Irish data, drawing from the de-duplicated dataset released by the UCCIX team \cite{tran2024uccixirishexcellencelargelanguage}. Additionally, a subset of the National Corpus of Irish was made available for this project under the CC BY-SA 4.0 license. Due to copyright restrictions, the corpus was shuffled at the sentence level and copyright-protected data was not shared. An overview of the pre-training data and sources is shown in Table \ref{tab:data-sources}.

For English text, we used the first 50K samples from Wikipedia dump 20220301 \footnote{https://dumps.wikimedia.org}, to cover a diverse range of topics. 
\begin{table}[t!]
\caption{Pre-training corpora and character counts. Note: Total character count is 3.265B.}
\label{tab:data-sources}
\centering
\footnotesize
\begin{tabular}{l r l r}
\toprule
\textbf{Source} & \textbf{Characters} & \textbf{Lang.} & \textbf{Prop.} \\
\midrule
The Bible          & 5M   & gle    & 0.0015 \\
UCCIX\_Leipzig     & 13M  & gle    & 0.0040 \\
UCCIX\_ELRC        & 17M  & gle \& eng & 0.0052 \\
UCCIX\_Gawiki      & 25M  & gle    & 0.0077 \\
UCCIX\_Gaparacrawl & 107M & gle    & 0.0328 \\
CNG                & 549M & gle    & 0.1681 \\
UCCIX\_Glot500     & 530M & gle    & 0.1623 \\
Wikipedia          & 819M & eng    & 0.2508 \\
UCCIX\_CulturaX    & 1.2B & gle    & 0.3675 \\
\midrule
\textbf{Total}     & \textbf{3.265B} &       & 1.0000 \\
\bottomrule
\end{tabular}
\end{table}

\subsubsection{De-duplication}
Near duplicate documents introduce redundancy in training. In the context of this project, the data released by the UCCIX team has already undergone pre-processing to remove duplicates. Therefore, our most pertinent question was whether the CNG and Bible data were already \textit{contained} \cite{containment_broder} in the UCCIX data. Containment (MinHash) is defined below in Equation \ref{containment_eq}, where $|A| < |B|$. To measure the containment, the text was lower-cased and punctuation was removed. Documents were transformed (L) into sets of unique 5-gram deterministic hashes in line with UCCIX pre-processing methodology. \cite{tran2024uccixirishexcellencelargelanguage}.
\begin{equation}
\label{containment_eq}
    \frac{|L(A)\cap L(B)|}{|L(A)|}
\end{equation}
 The containment of CNG in the UCCIX was low, whereas much of The Bible (>60\%) was contained in the UCCIX corpus. We did not remove the Bible from the pre-training mixture due to its long-context, high quality, and small size.

\subsubsection{Segmentation}
The end of document token \texttt{<|endoftext|>} \cite{QwenConcepts} was inserted between samples across all data sources, preventing the model from inferring dependencies between unrelated documents and leveraging a special token it is already trained with. For the Dáil dataset, this corresponded to between speaker utterances; for CNG, the end of text token was appended to each sentence (given the sentence-level data shuffling of this dataset); and for the UCCIX dataset between each textual sample in the dataset. Each parallel eng-gle sample was explicitly prepended with its respective language tag: "[eng]"/"[ga]".

\subsubsection{Evaluation and Baselines}
We quantify the model's performance with the same benchmarks as UCCIX. These benchmarks evaluate both Irish and English language skills across closed questions, translation tasks, topic identification, and Irish grammar. In order to measure the relative performance of Qomhrá, we compare against two models of the same size, one having undergone no CPT for Irish, Llama 3.1-8B, and the other having been adapted to Irish via CPT, UCCIX. 
To understand training dynamics, we evaluate the base model, the model after one epoch and the model after two epochs of CPT. This measures the impact of CPT on the model's performance and the extent to which further epochs improve adaption.  

\subsection{Continued Pre-training}
For low-resource languages, the adaption of an existing LLM avoids training from scratch. This draws from the base model's existing linguistic understanding to reduce training overhead. We explore various adaptation methods and their suitability to developing Qomhrá. 

While methods like in-context learning \cite{ICL_cahyawijaya-etal-2024-llms} and parameter efficient fine-tuning (PEFT) \cite{LoRA_2021, QLoRA} offer efficient adaptation, continued pre-training (CPT) \cite{gururangan-etal-2020-CPT} is better suited for the large domain shift of a new language \cite{Lu_Luu_Buehler_2025_CPT_LORA}.

\subsubsection{Training Configuration and Hyper-parameters}
For the CPT configuration, we pack the text into 2048-token blocks, significantly reducing the Qwen3-8B maximum context window of 128K due to memory constraints. We split the pre-training data 94:3:3 (train:validation:test) to monitor training stability. In line with UCCIX methodology, we prepend bitext to smoothen domain shift, before mixing and  shuffling the monolingual English and Irish data. Shuffling is done  deterministically for reproducibility. 

A per-device batch size of one with gradient accumulation ($\times8$) was used to stay within memory constraints. DeepSpeed ZeRO-2 enabled data parallelism with both optimiser and gradient partitioning across GPUs. Gradient checkpointing reduced memory overhead. Training ran for two epochs to balance convergence with constraints, with validation monitoring from Weights \& Biases \cite{wandb} and a test script to load model from checkpoint and run test generation every 3K steps to assess progress.

The AdamW optimiser was used along with the default hyper-parameters of $\beta_1 = 0.9$, $\beta_2 = 0.999$, $\epsilon=1e^{-8}$ and a learning rate of $1 e^{-4}$.

The BF16 \cite{google-cloud-bfloat16} floating-point format was used as it approximates the dynamic range of FP32 with the 50\% memory reduction FP16 provides \cite{kalamkar2019stud_BF16y}, which is important to maximise compute resources while maintaining training stability.

\subsection{Instruction Tuning}
Instruction tuning transitions an LLM from a token predictor to a chatbot, useful for interfacing with humans. Qomhrá is trained on instruction-response pairs which requires significantly less data \cite{wei2022FLAN}. We translate the Dolly V2 \cite{DatabricksBlog2023DollyV2} dataset to Irish. A translation of this dataset contributes a parallel Irish and English instruction-tuning dataset. The quality, however, is bound by the machine translation model. As such, an experiment was set up to determine the strongest closed-weight LLM for Irish, outlined in Section 4.1.

The strongest model, as determined by this experiment, went on to translate the Dolly V2 instruction-tuning dataset. Qomhrá is then fine-tuned on the dataset using Low-Rank Adaptation (LoRA) to create Qomhrá-Instruct. Qomhrá-Instruct is then re-benchmarked on the same benchmarks as Qomhrá to evaluate the impact of the dataset.

\subsubsection{Training Configuration and Hyper-parameters}
The Qwen chat template was applied to prompts. The thinking tags were removed as the goal was not to develop a reasoning model. BF16 format was used and learning rate hyperparameters were tuned in line with the Unsloth LoRA Hyper-parameter Guide \cite{unsloth2025lora}. The learning rate was $2e^{-4}$ with AdamW optimiser, rank 16, $\alpha=32$, weight decay $=0.01$ and 3\% warm up ratio for fast adaptation.

\subsection{Human Feedback}
We propose a novel method that synthesises ``accepted'' and ``rejected'' responses, to allow for direct preference optimisation without any human annotation. This facilitates alignment in extremely low-resource settings, where access to L1 speakers is limited and costly. We instructed Gemini-2.5-Pro to generate paired translations for each English instruction: one high-quality ("accepted") and one low-quality ("rejected")
 
To do this, Gemini-2.5-Pro was instructed to translate each English instruction response sample, where the prompt is identical but the response varies, where an L1 speaker is expected to ``accept'' one response and ``reject'' the other. The LIMA dataset \cite{LIMA} was selected for translation, due to its effectiveness at small scale. The top-ranked model, Gemini-2.5-Pro, was used to synthesise the preference dataset. The following is the prompt we designed, specifying high contrast and strict JSON formatting for robust API parsing. 
\begin{quote}
{\ttfamily
Translate the following English Instruction and response into Irish.\\
- response1 should be a natural, direct and fluent translation,\\
- response2 should be a weak alternative, it should be  unhelpful, not idiomatic, inaccurate, awkward.\\
- The contrast in quality of Irish should be very high.\\
OUTPUT FORMAT (STRICT):\\
Return strict JSON with exactly:\\
\{\\
"instruction": "$<$instruction in Irish$>$",\\
"response1": "$<$much better response in Irish$>$",\\
"response2": "$<$much worse response in Irish$>$"\\
\}\\
The following is the English prompt-response pair:
}
\end{quote}

The English instruct-response pair was appended to the request.

A validation experiment was set up where the L1 speaker indicated preference between response A and response B. The in-pair ordering was randomly shuffled and the original model-intended preference was stored with each annotation. This experiment determined whether the L1 speaker agreed with the LLM's prescription of high-quality and low-quality translations.

\section{Experimental Setup and Evaluation}

\subsection{Translator Selection Methodology}
The top three commercial LLM providers were selected: OpenAI, Google, and Anthropic. For each provider, both the most recent flagship model and a less expensive model were tested, shown in Table \ref{tab:Closed_LLMs_table}. The budget model was considered as these model providers do not specifically evaluate Irish language capabilities and thus the flagship model is not necessarily the best or most cost-effective model. Therefore, we compare flagship vs budget performance which is particularly relevant for low-resource scenario data synthesis.

\begin{table}[b!]
\centering
\caption{Models evaluated for Irish language translation generation.}
\label{tab:Closed_LLMs_table}
\begin{tabular}{@{}lll@{}}
\toprule
\textbf{Provider} & \textbf{Flagship Model} & \textbf{Budget Model} \\
\midrule
OpenAI    & GPT-5              & GPT-5-mini         \\
Anthropic & Claude-4-Sonnet    & Claude-3.5-Haiku   \\
Google    & Gemini-2.5-Pro     & Gemini-2.5-Flash   \\
\bottomrule
\end{tabular}
\end{table}

Each model was provided with Irish language text from the Irish parliament (Dáil) and Wikipedia and asked to generate a prompt-response pair. This follows methodology applied to generate synthetic Kazakh instruction-tuning data \cite{laiyk-etal-2025-instruction}. 

120 annotations were completed by both the L2 and the L1 speaker, whereas the LLMs annotated all 600 samples. The ranking model selected was the Bradley-Terry model due to its strong performance dealing with small sample sizes \cite{daynauth-etal-2025-ranking}.

\subsection{Translator Selection Analysis}
Table \ref{tab:aggregated_raters} shows that both the L2 and the L1 speaker were aligned in evaluating Gemini-2.5-Pro as the strongest model for Irish language text generation. Therefore, Gemini-2.5-Pro was used for subsequent synthetic dataset creation as the L1 speaker is considered the gold standard.

Claude, Gemini, and Chat-GPT ranked a GPT model as the number one model for Irish language text generation. A plausible hypothesis for this is that these models have seen GPT-generated outputs and web content, inducing bias in favour of GPT-style outputs based on familiarity. With moderate differences of inter-LLM agreement, it is difficult to discern a pattern, but this trend could indicate model-specific biases which would be interesting to explore further. 

\begin{table}[t!]
\centering
\small 
\caption{Bradley-Terry ranking of models. Abbreviations: 
GEM-Pro (Gemini-2.5-Pro), Claude-Sonnet (Claude-4-Sonnet), 
Claude-Haiku (Claude-3.5 Haiku), GEM-Flash (Gemini-2.5-Flash).}
\label{tab:aggregated_raters}
\setlength{\tabcolsep}{4pt} 
\begin{tabular}{@{}l lll@{}}
\toprule
Rank & L1 Speaker & L2 Speaker & LLMs. \\
\midrule
1 & GEM-Pro     & GEM-Pro       & GPT-5        \\
2 & Claude-Sonnet & GPT-5-mini    & GPT-5-mini   \\
3 & GPT-5       & Claude-Haiku  & GEM-Pro      \\
4 & Claude-Haiku& GEM-Flash     & Claude-Sonnet\\
5 & GEM-Flash   & GPT-5         & GEM-Flash    \\
6 & GPT-5-mini  & Claude-Sonnet & Claude-Haiku \\
\bottomrule
\end{tabular}
\end{table}

\subsection{Pre-training}
Pre-training was carried out with two Nvidia H100 GPUs with 80GB VRAM for 34,360 steps with a total train-time of 44.196 hours. The benchmarks are displayed in Table \ref{tab:pretrain-benchmark}.

\begin{table*}[t!]
\caption{Pre-training \& Instruction-Tuning Benchmarking results}
\label{tab:pretrain-benchmark}
\centering
\setlength{\tabcolsep}{4pt}
\begin{tabular}{l|ccccccc}
\hline
Model & Cloze-gle & SIB-gle & IQA-gle & IQA-eng & BLEU eng2gle & BLEU gle2eng & NQ-eng \\
\hline
Llama-3.1-8B   & 0.59 & 0.7696 & 0.4861 & 0.7747 & 0.0880 & 0.4229 & \textbf{0.2767} \\
UCCIX          & 0.75 & 0.7794 & 0.3889 & 0.3704 & \textbf{0.3334} & \textbf{0.4636} & 0.1668 \\
Qwen3-8B-Base  & 0.44 & 0.6471 & 0.4633 & 0.8025 & 0.0154 & 0.2684 & 0.2590 \\
Qomhrá-1e-CPT  & 0.85 & \underline{\textbf{0.8529}} & \underline{\textbf{0.6810}} & \underline{\textbf{0.8177}} & 0.0368 & 0.0509 & 0.0374 \\
Qomhrá-2e-CPT  & 0.86 & 0.8480 & \underline{\textbf{0.6810}} & 0.8025 & 0.0363 & 0.0519 & 0.0355 \\
Qomhrá-Instruct & \underline{\textbf{0.88}} & 0.8186& 0.6760 & 0.7924 & \underline{0.1167}& \underline{0.0770}  & \underline{0.1269} \\
\hline
\end{tabular}
\par\medskip
\begin{minipage}{0.85\linewidth}
\footnotesize
\textbf{Bold} indicates the best performing model whereas \underline{underline} indicates the best performing Qomhrá model. 1e and 2e refer to 1 and 2 epochs of CPT.
\end{minipage}
\end{table*}

\subsubsection{Benchmark Descriptions}
To provide clarity on the evaluations, we provide a detailed description of the five benchmarks used as there are none currently available.

\textbf{Cloze-gle} \cite{tran2024uccixirishexcellencelargelanguage} evaluates the model's grasp of Irish grammatical gender and person agreement (masculine/feminine/plural). The model is presented with three candidate sentences differing only by a single pronoun (e.g., \textit{é, í, iad}). The log-likelihood (LL) of each full sentence is computed. Since the candidates differ by only one word, the sentence with the highest likelihood represents the model's prediction.
In Table \ref{tab:cloze_example}, the noun \textit{faoiseamh} is masculine. The model assigns the highest likelihood (-38.2) to the sentence containing the masculine pronoun \textbf{é}, matching the target.

\begin{table}[H]
\centering
\small
\resizebox{\linewidth}{!}{%
\begin{tabular}{@{}llcc@{}}
\toprule
\textbf{\#} & \textbf{Candidate Sentence} & \textbf{LL} & \textbf{Target} \\
\midrule
1 & Is iontach an faoiseamh \textbf{í} sin. & -43.4 & No \\
2 & Is iontach an faoiseamh \textbf{é} sin. & \textbf{-38.2} & \textbf{Yes} \\
3 & Is iontach an faoiseamh \textbf{iad} sin. & -43.4 & No \\
\bottomrule
\end{tabular}%
}
\caption{Cloze-gle Example. The distinguishing pronoun is bolded.}
\label{tab:cloze_example}
\end{table}

\textbf{SIB-gle} \cite{adelani2024sib200simpleinclusivebig} tests topic modelling capabilities. Unlike the zero-shot Cloze task, this is a 10-shot evaluation. The model is presented with input text and must assign it to one of seven high-level topics (e.g., \textit{eolaíocht/teicneolaíocht} [science/tech], \textit{polaitíocht} [politics]).
The model scores each topic label as a continuation; the label with the highest probability is selected. In Table \ref{tab:sib_example}, the text describes how a nuclear bomb is made and the model correctly assigns the highest probability to the science category. Only 2 of the 7 categories are displayed in Table \ref{tab:sib_example} for brevity.

\begin{table}[H]
\centering
\small
\resizebox{\linewidth}{!}{%
\begin{tabular}{@{}llcc@{}}
\toprule
\textbf{Text} & \multicolumn{3}{p{0.75\linewidth}}{\textit{Oibríonn an chéad bhuama eamhnaithe ar an bprionsabal gur gá fuinneamh...}} \\
\midrule
\textbf{\#} & \textbf{Candidate Label} & \textbf{LL} & \textbf{Target} \\
\midrule
1 & eolaíocht/teicneolaíocht & \textbf{-0.017} & \textbf{Yes} \\
2 & polaitíocht & -6.264 & No \\
\bottomrule
\end{tabular}%
}
\caption{SIB-gle Example (truncated candidates).}
\label{tab:sib_example}
\end{table}

\textbf{IQA-gle \& IQA-eng} \cite{tran2024uccixirishexcellencelargelanguage}
assess multiple-choice question answering in both English and Irish using a 5-shot context. For a given question, the model is presented with four candidate answers. The log-likelihood of each candidate answer is calculated given the question context.
Table \ref{tab:qa_example} illustrates an example of IQA-eng where the model correctly identifies that Irish is taught as a compulsory subject. The method is the same for IQA-gle.

\begin{table}[H]
\centering
\small
\resizebox{\linewidth}{!}{%
\begin{tabular}{@{}llcc@{}}
\toprule
\textbf{Q} & \multicolumn{3}{l}{How is the Irish language taught in secondary education?} \\
\midrule
\textbf{\#} & \textbf{Candidate Answer} & \textbf{LL} & \textbf{Target} \\
\midrule
1 & As an optional subject & -1.7 & No \\
2 & As an extracurricular activity & -4.0 & No \\
3 & As a compulsory subject & \textbf{-0.3} & \textbf{Yes} \\
4 & As a foreign language & -3.2 & No \\
\bottomrule
\end{tabular}%
}
\caption{IQA-eng Example.}
\label{tab:qa_example}
\end{table}

\textbf{BLEU (gle $\leftrightarrow$ eng)} \cite{lankford-etal-2022-gahealth}
evaluates translation quality via BLEU-4 in a 5-shot setting on health data. BLEU-4 measures the n-gram overlap (up to 4-grams) between the model's response and a gold-standard reference.
Table \ref{tab:bleu4_example} shows a sample generation where Qomhrá's response omits the definitive plural article \textit{na}, this difference is highlighted in bold.

\begin{table}[H]
\centering
\small
\resizebox{\linewidth}{!}{%
\begin{tabular}{@{}p{0.25\linewidth} p{0.75\linewidth}@{}}
\toprule
\textbf{Source (en)} & Important notice: Latest information on Revenue services and tax and customs measures... \\
\midrule
\textbf{Reference (gle)} & Fógra tábhachtach: An t-eolas is déanaí faoi sheirbhísí na gCoimisinéirí Ioncaim agus \textbf{na} bearta cánach agus custam... \\
\midrule
\textbf{Response (gle)} & Fógra tábhachtach: An t-eolas is déanaí faoi sheirbhísí na gCoimisinéirí Ioncaim agus bearta cánach agus custam...\\
\bottomrule
\end{tabular}%
}
\caption{BLEU gle2eng Example}
\label{tab:bleu4_example}
\end{table}

\textbf{NQ-eng} \cite{nq-2019} evaluates world knowledge via open-ended generation (5-shot). Success is measured by an exact match: the generated text must be a sub-string of a valid reference answer.
Table \ref{tab:nq_example} demonstrates a failure case. Although the model correctly identifies the date ("October 24"), it fails to specify the year required by the reference. This strict metric explains the lower scores observed in Table \ref{tab:pretrain-benchmark}.

\begin{table}[H]
\centering
\small
\resizebox{\linewidth}{!}{%
\begin{tabular}{@{}ll@{}}
\toprule
\textbf{Question} & When did Taylor Swift's first album release? \\
\midrule
\textbf{Response} & October 24 \\
\textbf{Reference} & ["October 24, 2006", "2005"] \\
\textbf{Exact Match} & \textbf{No} \\
\bottomrule
\end{tabular}%
}
\caption{NQ-eng Example (Exact Match failure).}
\label{tab:nq_example}
\end{table}

\subsubsection{Benchmark Results}
The success on both English and Irish benchmarks after CPT demonstrates the effectiveness of bilingual pre-training. Qomhrá outperforms the other two models in the Cloze benchmark closely followed by UCCIX. Its higher performance compared to Llama-3.1-8B is unsurprising as the gender information of vocabulary words can only be learned through exposure to the language.

Qomhrá outperformed UCCIX by 29.2\% in Irish as opposed to 44.4\% in English on the IQA benchmark. The information from these questions is the same so language capability must be the factor that caused Qomhrá to outperform UCCIX, indicating improved English performance retention for Qomhrá, likely due to its higher proportion of English data at 25\% compared to UCCIX's 1\% . Qomhrá did not improve performance with the second epoch of training. This is in line with expectations that base models saturate after one epoch of CPT \cite{Lu_Luu_Buehler_2025_CPT_LORA}. Therefore in future development of the model, CPT would only be trained for one epoch to prevent redundant compute expenditure.

\subsection{Instruction Tuning}
Qomhrá only has lower accuracy than UCCIX in Table \ref{tab:pretrain-benchmark} on \textbf{open-ended} generation benchmarks: BLEU (eng2gle \& gle2eng) and NQ. This can be partly explained by the length of Qomhrá's responses. Following instruction tuning, Qomhrá demonstrates improved performance as the model learns to stop generating tokens once the user's question is answered.

To quantify this, we compare the response-length distributions before and after instruction tuning. We use a non-parametric directional Mann-Whitney U test as distributions are not ostensibly normal. The base model has longer responses than the instruct model across all three open-ended benchmarks ($p<0.001$). Noticeably, the base model always generated longer sequences than the instruct model ($p=0$) for the NQ benchmark indicating zero overlap in response length distributions. 

This improved performance after instruction tuning was unsurprising as its intention is to improve unseen task performance \cite{wei2022FLAN}. Future work could extend instruction tuning by adding cultural knowledge through the translation of parliamentary and government data as seen to be effective for Kazakh \cite{laiyk-etal-2025-instruction}.

\subsection{Human Feedback}
After we generated the synthetic human preference data with Gemini-2.5-Pro, the L1 speaker annotated 91 samples of this data. From these 91 annotations, 90/91 were aligned, giving a Cohen's $\kappa$ of 0.978, indicating near-perfect alignment. This suggests that all 1K synthetic human preference samples are L1-speaker aligned, allowing the dataset scaling required for training. 

Contrary to the instruction tuning inter-annotator agreement, the LLM and the human showed near-perfect annotation alignment. We hypothesise that this reflects the level of contrast between candidates in the annotations, i.e., LLMs can be used to substitute human annotators in low-resource languages for less complex tasks. 

\section{Conclusion}
Our paper presents Qomhrá, a bilingual Irish and English LLM, outlining a full CPT and instruction-tuning pipeline with human preference data synthesis. Our increased English language proportion in the CPT training highlights its necessity when training systems adapted to real-world context, where the low-resource and high-resource language co-exist. 

We provide the first human evaluation comparison of existing LLMs for Irish. This guides anyone seeking to integrate LLMs into language technology for Irish. In creating a 30K parallel instruction dataset and a 1K human preference dataset, we contribute a validated labelled data at a scale that can assist others in developing Irish-language LLMs. 

Furthermore, our inter-annotator analysis showed that while L2 speakers and LLMs failed to align with an L1 speaker for complex annotation, an LLM can create a high-quality DPO dataset ($\kappa=0.978$ alignment), if candidate contrast is high. This provides a useful heuristic for others considering data synthesis using existing LLMs to distill knowledge. 

Our results establish key elements useful for developing chatbots for Irish and we hope that work of this nature can spur and aid others in developing LLMs for Irish and other low-resource languages. 

\subsection*{Limitations}
Though the closed-weight LLMs evaluated at the time of research were the state of the art, model providers have released new iterations since then, motivating updated comparisons. Due to compute limitations, our work is limited to the 8B model scale and a focus on Irish and English. Future work should include more languages and base models to support the generalization power of results. 

Equally, only one open-weight model: Qwen3 served as the base model, where in future, we recommend independent empirical testing of available open-weight models specific to domain adaption, i.e., evaluating a proxy for Irish language performance before commencing the costly CPT, fine-tuning pipeline. 

Regarding our annotator alignment findings, they are limited by access to the inclusion of only one L1 speaker, which encourages future work to collaborate at a larger scale with the low-resource language community.  

Furthermore, evaluation does not address cultural alignment, code-switching, or dialectal variation, which are key functionalities that should be addressed in future work. 

\subsection*{Ethical Considerations}
Developing human language technology imposes ethical considerations concerning the human interfacing with the technology. As Qomhrá is primarily trained on web-crawled data, content can potentially be harmful. Future research should involve more rigorous pre-processing to remove this content. Qomhrá is primarily released for research purposes and users of Qomhrá should be mindful of potential harms in its generated outputs.  

LLM CPT requires significant compute power which contributes to greenhouse gas emissions, as such it is important to quantify our impact. The power usage of an Nvidia H100 GPU is 700W. We trained for 44.196 hours on 2 GPUs. The carbon intensity was approximately 200 gCO$_2$/kWh \cite{CI-Eirgrid}, which gives a total of 12.37kg CO$_2$. This can be understood relative to the average of 10.4 tonnes of CO$_2$ per capita in Ireland \cite{CSO-Env}. Efficient corpus sampling methods can be used in future work for more efficient training.
\bibliographystyle{acl_natbib}
\bibliography{acl_latex}

\end{document}